\documentclass[conference]{IEEEtran}
\IEEEoverridecommandlockouts
\usepackage{cite}
\usepackage{amsmath,amssymb,amsfonts}
\usepackage{algorithmic}
\usepackage{authblk}
\usepackage{graphicx}
\usepackage{textcomp}
\usepackage{subcaption}
\usepackage{array}
\usepackage{xcolor}

\usepackage[english]{babel}
\usepackage{lipsum}
\usepackage[pscoord]{eso-pic}

\def\BibTeX{{\rm B\kern-.05em{\sc i\kern-.025em b}\kern-.08em
    T\kern-.1667em\lower.7ex\hbox{E}\kern-.125emX}}
\begin{document}

\title{Cross Lingual Speech Emotion Recognition: Urdu vs. Western Languages}

\author[1]{Siddique Latif}
\author[1]{Adnan Qayyum}
\author[2]{Muhammad Usman}
\author[1]{Junaid Qadir}

\affil[1]{Information Technology University (ITU)-Punjab, Pakistan}
\affil[2]{COMSATS University Islamabad (CUI), Islamabad}

\maketitle

\begin{abstract}
Cross-lingual speech emotion recognition is an important task for practical applications. The performance of automatic speech emotion recognition systems degrades in cross-corpus scenarios, particularly in scenarios involving multiple languages or a previously unseen language such as Urdu for which limited or no data is available. In this study, we investigate the problem of cross-lingual emotion recognition for Urdu language and contribute URDU---the first ever spontaneous Urdu-language speech emotion database. Evaluations are performed using three different Western languages against Urdu and experimental results on different possible scenarios suggest various interesting aspects for designing more adaptive emotion recognition system for such limited languages. In results, selecting training instances of multiple languages can deliver comparable results to baseline and augmentation a fraction of testing language data while training can help to boost accuracy for speech emotion recognition. URDU data is publicly available for further research\footnote{https://github.com/siddiquelatif/URDU-Dataset}. 
\end{abstract}


\section{Introduction}

Machine learning models are playing important roles in different speech and audio based intelligent system for effective human-computer interaction \cite{latif2018phonocardiographic,qayyum2018quran}. Recently, speech emotion recognition has attained substantial interest in recent years due to its various potential applications in robotics, business, education, healthcare to name a few. Researchers have addressed the emotional state identification problem with different cutting-edge machine learning models and using different speech features as input to the system \cite{latif2017variational,latif2018adversarial}. These systems perform well when tested on the same corpus but much more poorly when performance is tested on speech utterances from different corpus or language \cite{latif2018cross}. Cross-corpus speech emotion recognition is a challenging task due to dissimilarities and sparseness among databases. For example, the difference in corpus language, speakers' age, labeling schemes, and recording conditions considerably affect the models' performance when tested across different databases \cite{zhang2016cross,zhang2011unsupervised}. This motivates the design of more robust emotion recognition systems that can detect emotion from cross-corpus data with multiple languages. This will enable the real-time deployment of speech emotion recognition systems for a wide verity of health and industry related applications.  

In automatic speech emotion detection, most of the studies has focused on a single corpus, without considering cross-language and cross-corpus effects. One of the reason is that we have only a few numbers of corpora for research on speech analysis as compared to the number of spoken languages in the world \cite{wang2015transfer}. Additionally, available resources are highly diverse even when considering just English language, which leads to the data sparsity problem noted in speech emotion recognition research \cite{latif2018transfer}. In such situations, it is impossible for automatic speech emotion recognition system to work in real-time on every language by learning from single data resource. Therefore, for practical applications, we need more adapted models that can learn from multiple resources in different languages.   

Cross-corpus emotion recognition has been studied by various researchers to improve the classification accuracy across different languages. These studies used various publicly available datasets to highlight the interesting trends in cross-corpus emotion recognition \cite{schuller2010cross,eyben2010cross}. Although, a few studies have address cross-corpus emotion recognition problem as reported in \cite{schuller2010cross}, issues for emotion recognition for minority languages like Urdu have not been explored. Urdu is the official national language of Pakistan and it is one of the 22 official languages recognized in the Constitution of India\footnote{https://en.wikipedia.org/wiki/Urdu}. Dealing with such languages is very crucial and necessary for the functionality of next-generation systems \cite{albornoz2017emotion} to adapt information from multiple languages for this task. Similar trends to this case are highlighted in this work.

In this paper, we propose an emotion recognition system to model the Urdu language. In our previous work \cite{latif2018cross}, we studied cross-corpus emotion recognition using five different datasets of Western languages and showed how accuracy for emotion detection can be improved. For this study, we formulated the first spontaneous emotional dataset in Urdu language that contains authentic emotional speech and is publicly available. To evaluate this dataset, we experimented different possible scenarios for within- and cross-corpus emotion recognition using three other Western languages (English, Italian, and German) databases and showed how emotion recognition across different languages can be achieved using data of different languages. To best of our knowledge, this is the first study on Urdu language that presents the performance trends of automatic emotion recognition system using different Western languages. The findings in this study would be very helpful for designing speech emotion recognition system for practical applications not only in Pakistan and various other countries. 

The rest of the paper is organized as follows. In the next section, we present background and related work. In section \ref{meth}, we discuss proposed schemes and our methodology. In section \ref{exp}, we present the experimental procedure and results followed by the discussion of results in section \ref{dis}. Finally, we conclude the paper in section \ref{con}.

\section{Related Work}

Cross-language emotion detection from speech using machine learning techniques has been studied in different research works. These studies have mostly pointed out the need for in-depth research by performing some preliminary empirical analysis on cross-corpus emotions learning. For instance, Elbarougy et al. \cite{elbarougy2014toward} have investigated the differences and commonalities of emotions in valence-activation space among three languages (Japanese, Chinese, and  German) by using thirty subjects and proved that emotions are similar among subjects speaking different languages. Schuller et al. \cite{schuller2010cross} have performed evaluation experiments on cross-corpus emotion recognition using SVM on six corpora in three languages (English, Danish, and German) and suggested the need of future research for the limitations of current systems of emotion recognition. Eyben et al. \cite{eyben2010cross} evaluated some pilot experiments using SVM on four corpora of two languages (English and German) to show the feasibility of cross-corpus emotion recognition. In \cite{parlak2014cross}, authors introduced EmoSTAR as new emotional corpus and performed cross-corpus tests with EMO-DB using SVM. 

To study the universality of emotional cues among languages, Xia et al. \cite{xiao2016speech} studied speech emotion recognition for Mandarin vs. Western languages (i.e., German, and Danish). They evaluated gender-specific speech emotion classification and achieved the accuracy more than the chance level. In \cite{schuller2015cross}, authors used six emotional databases and evaluated different scenarios for cross-corpus speech emotion recognition. They were able to capture the limitations of current systems due to their very poor performance on spontaneous set or natural emotional corpus. In another interesting work, Albornoz et al. \cite{albornoz2017emotion} developed an emotion profile based ensemble SVM for emotion recognition in different unseen languages. The authors used the RML emotion database that covers six languages. However, this data is recorded under the same condition and contains very small number of utterances, five sentences, for each emotion. 

In this study, we consider the development of Urdu language data to empirically evaluate cross-lingual emotion recognition using three Western languages. We experimented with different possible scenarios for cross-lingual emotion recognition to highlight the limitations of speech emotion recognition system for Urdu language and provide different interesting insights for designing an adaptive speech emotion recognition system. This study not only has great commercial value in Pakistan but also for next-generation adaptive mood inference engines to cope with logistic and cultural differences.

\section{Methodology}
\label{meth}
We have selected EMO-DB, SAVEE, EMOVO datasets and one in Urdu language called URDU that is formulated for this task. The datasets are selected to incorporate maximum diversity in the languages. The chosen set of databases also provide a good combination of acted (SAVEE, EMOVO, and EM-ODB) and natural emotion (URDU) to evaluate cross-corpus emotion detection. Each corpus has different annotations and recorded in different studio conditions. For any classification experiment, it is necessary that training and testing must have same class labels. Therefore, we consider binary (positive/negative) valance (see Table \ref{table: MAP}) mapping as used in \cite{schuller2010cross,deng2013sparse,eyben2016geneva}.

The further details on the selected databases, speech features, and classifier are presented below. 

\begin{table*}[ht]
\centering
\caption{Selected corpora information and the mapping of class labels onto Negative/Positive valence.}
\begin{tabular}{|m{1.2cm}|m{1.3cm}|m{1cm}|m{1.3cm}|m{4.7cm}|m{3.28cm}|m{1.4cm}|}
\hline
\textbf{Corpus}
&\textbf{Language}
&\textbf{Age}
&\textbf{Utterances}
&\textbf{Negative Valance}
&\textbf{Positive Valance}
&\textbf{References}
\\ \hline

 \begin{tabular}[c]{@{}l@{}}EMO-DB\end{tabular}
&\begin{tabular}[c]{@{}l@{}}German\end{tabular}
&\begin{tabular}[c]{@{}l@{}}Adults\end{tabular}
&\begin{tabular}[c]{@{}l@{}}494\end{tabular}
&\begin{tabular}[c]{@{}l@{}}Anger, Sadness, Fear, Disgust, Boredom\end{tabular} 
&\begin{tabular}[c]{@{}l@{}}Neutral, Happiness\end{tabular}
&\begin{tabular}[c]{@{}l@{}}\cite{burkhardt2005database}\end{tabular}
\\ \hline
 \begin{tabular}[c]{@{}l@{}}SAVEE\end{tabular}
&\begin{tabular}[c]{@{}l@{}}English\end{tabular}
&\begin{tabular}[c]{@{}l@{}}Adults\end{tabular}
&\begin{tabular}[c]{@{}l@{}}480\end{tabular}
&\begin{tabular}[c]{@{}l@{}}Anger, Sadness, Fear, Disgust \end{tabular}
&\begin{tabular}[c]{@{}l@{}}Neutral, Happiness, Surprise\end{tabular}
&\begin{tabular}[c]{@{}l@{}}\cite{jackson2014surrey}\end{tabular}
\\ \hline 
 \begin{tabular}[c]{@{}l@{}}EMOVO\end{tabular}
&\begin{tabular}[c]{@{}l@{}}Italian\end{tabular}
&\begin{tabular}[c]{@{}l@{}}Adults\end{tabular}
&\begin{tabular}[c]{@{}l@{}}588\end{tabular}
&\begin{tabular}[c]{@{}l@{}}Anger, Sadness, Fear, Disgust\end{tabular}
&\begin{tabular}[c]{@{}l@{}}Neutral, Joy, Surprise\end{tabular}
&\begin{tabular}[c]{@{}l@{}}\cite{costantini2014emovo}\end{tabular}
\\ \hline
 \begin{tabular}[c]{@{}l@{}}URDU\end{tabular}
&\begin{tabular}[c]{@{}l@{}}Urdu\end{tabular}
&\begin{tabular}[c]{@{}l@{}}Adults\end{tabular}
&\begin{tabular}[c]{@{}l@{}}400\end{tabular}
&\begin{tabular}[c]{@{}l@{}}Angry, Sad\end{tabular}
&\begin{tabular}[c]{@{}l@{}}Neutral, Happy\end{tabular}
&\begin{tabular}[c]{@{}l@{}}Author\end{tabular}
\\ \hline

\end{tabular}
\centering
\label{table: MAP}
\end{table*}

\subsection{Speech Databases}

\subsubsection{EMO-DB}
It is a well known and widely used speech corpus for automatic emotion classification. The language of recordings of this dataset is German and it was introduced by \cite{burkhardt2005database}. It contains utterances by ten professional actors in anger, boredom, disgust, fear, joy, neutrality, and sadness. The linguistic content in this data is pre-defined in emotionally neutral ten German short sentences. The whole dataset contains over 700 utterances, while only 494 phrases are emotionally labeled that we used in this study. 

\subsubsection{SAVEE}
Surrey Audio-Visual Expressed Emotion (SAVEE) database \cite{jackson2014surrey} is another famous multimodal corpus that contains the recordings from 4 male actors in 7 different emotions. This dataset includes 480 British English utterances in total. The recordings were assessed by 10 evaluators under audio, visual, and audio-visual conditions to check the quality of performance by actors. The scripts used in this data were chosen from the standard TIMIT corpus \cite{garofolo1993darpa} and utterance are labeled on 7 emotions (neutral, happiness, sadness, anger, surprise, fear, and disgust). We used all these emotions in our study.

\subsubsection{EMOVO}
This is the first Italian language emotional corpus that contains recordings of 6 actors who acted 14 sentences to simulate 7 emotional states (disgust, fear, anger, joy, surprise, sadness and neutral). This corpus contains overall 588 utterances and annotated by two different groups of 24 annotators. All the recordings were made in the Fondazione Ugo Bordoni laboratories. 
\subsubsection{URDU}
We have collected the first custom dataset of spontaneous emotional speech in the Urdu language. The data consists of audio recordings collected from the Urdu TV talk shows. There are overall 400 utterances for four basic emotions: angry, happy, sad, and neutral. There are 38 speakers (27 males and 11 females).  This corpus contains spontaneous emotional excerpts from authentic and unscripted discussions between different guests of TV talk show. Vera-Am-Mittag \cite{grimm2008vera} corpus is an example of such a database, which is also collected from TV talk shows. In such databases, it is not possible to control the speakers' affective state that occurs in the cause of dialog. Therefore, it is always a trade-off between the naturalness of the interaction and controllability of affective content \cite{grimm2008vera}. 

In order to formulate this corpus, video clips are collected from YouTube based on the discussion and situations going on in the talk shows. Although emotional corpus formulation from TV shows is an easy and abundant task but obtaining samples in different emotions is very difficult. Sometimes the presence of music and other noise accompanies make the data collection worse. We cannot have all the emotional state compared to other databases that are recorded by experts in studio condition, therefore, we only collected videos on four basic emotions (i.e., happy, sad, angry, and neutral). After collecting video clips, we mixed and gave these files to four students from NUST\footnote{National University of Sciences and Technology (NUST), Islamabad, Pakistan} and CIIT\footnote{COMSATS Institute of Information Technology, Islamabad, Pakistan} universities. They were asked to annotate the data with emotional labels to the speakers' state using both audio and video content. Final labels were given to each utterance when at least 2 annotators assigned them same emotion. This dataset is publicly available for research purposes\footnote{https://github.com/siddiquelatif/URDU-Dataset}. 

\subsection{Feature Extraction}

In this study, we have used the recently proposed minimalistic parameters set called eGeMAPS \cite{eyben2016geneva}. These features are frame-level and knowledge-inspired features and have comparable results as compared to other popular speech features. The eGeMAPS features set includes Low-Level Descriptor (LLD) features that have been suggested as the most related to emotions by Paralinguistic studies \cite{eyben2016geneva}. Again, these features also have comparable and even better performance compared to large brute-force features while greatly reducing the feature dimensionality. Previously, they are proved to be useful for cross-corpus emotion recognition in \cite{latif2018cross}. The eGeMAPS consists of 88 features related to energy, frequency, cepstral, spectral, and dynamic information. The components of eGeMAPS selected from the arithmetic mean and coefficient of variation of 18 LLDs, 6 temporal features, 8 functionals applied to loudness and pitch, 4 statistics over the unvoiced segments, and 26 additional dynamic parameters and cepstral parameters. A list of these LLDs and functionals can be found in Section 3 of \cite{eyben2016geneva}.

For eGeMAPS feature extraction, we used openSMILE toolkit \cite{eyben2010opensmile}, which enables to extract a large number of audio features in stand alone as well as in real-time. This toolbox is written in C++ and is publicly available as both as a dynamic library and standalone command line executable file.

\subsection{Models for Classification}
To analyze cross-corpus speech emotion recognition, we use SVM that can provide robust classification results on data having a very large number of variables even with a small number of training examples \cite{vapnik2013nature}. Moreover, it can learn from very complex data and uses mathematical principles to handle overfitting problem as well \cite{razuri2015speech}. The simplicity and capability properties have made SVM very popular for classification problem, and it has been widely used for prototyping cross-corpus speech emotion recognition (for example \cite{schuller2010cross,eyben2010cross,deng2013sparse}).

SVM as a classifier constructs hyperplanes in a multidimensional space to separate samples of different class labels. It employs an iterative training algorithm to construct an optimal hyperplane that separates input data with a maximum margin hereby minimizing the following error function:
\begin{equation}
   \frac{1}{2} w^{T}w+C\sum_{i=1}^{N}\xi_{i}
\end{equation}
subject to the constraints:
\begin{equation}
    y_{i}\big(w^{T}\phi(x_{i})+b\big)\geq \xi_{i} \And{\xi_{i} \geq 0}, i=1,...,N
\end{equation}

where $C$ denotes the capacity constant that handles overfitting, $w$ represents the coefficients vector of, $b$ is a constant, $\xi_{i}$ shows the parameters to handle non-separable input data, and $i$ is the index labels of $N$ training samples. Note that $y\in\pm1$ represents the binary class labels and $x_{i}$ is the independent variables. $\phi$ is a kernel function that is used to transform input data to other feature space. 

For experiments, we map emotional classes of each corpus to binary valance dimension. eGeMAPS features are extracted from speech utterances and given to SVM that we used for binary emotion classification. SVM aims to construct optimal hyperplanes that separate the training data with maximum margin. The hyperparameters are selected using the validation data. We selected Gaussian kernel function due to its better performance as compared to the linear and cubic kernel on speech emotion classification during experimentations. 

\section{Experimental Setup and Results}
\label{exp}

In this section, we conduct three different experiments using SVM for cross-corpus speech emotion recognition to analyze the trends for Urdu and Western languages. The details of these experiments are given below.

\subsection{Baseline Results}

\begin{figure}[!ht]
\centering
\centerline{\includegraphics[trim=2cm 0.8cm 3.9cm 0.8cm,clip=true,width=.5\textwidth]{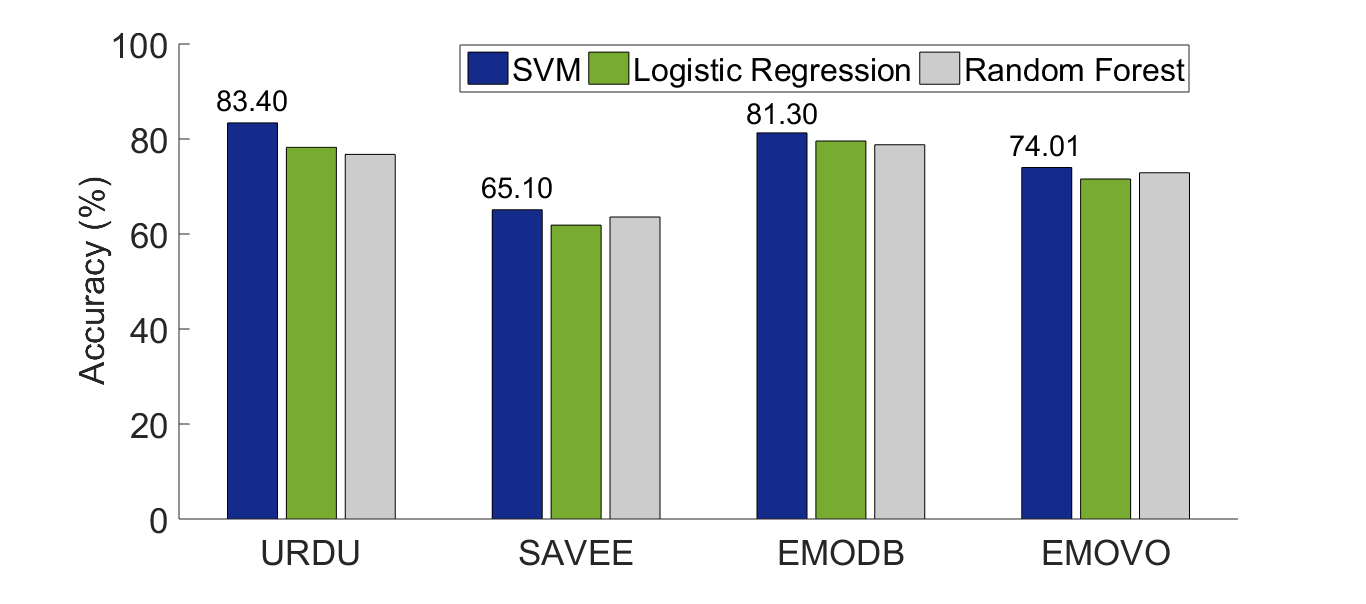}}
\caption{Baseline accuracy of different datasets using three classifiers.}
\label{Bls}
\end{figure}

In order to obtain baseline results, we performed classification experiments within a corpus using both training and testing data from the same corpus. The obtained results give us an idea about the best achievable accuracy within each corpus. We performed this experiment in speaker independent manner for all datasets and results are presented in terms of unweighted average recall rate (UAR). As URDU dataset consists of 38 speakers, so we selected 30 speakers for training and remaining 8 for testing with five-fold cross-validation. For others, we held one speaker out for testing and remaining speakers' data was used for training purpose with cross-validation equal to the number of speakers in the respective dataset. This is a most acceptable approach for speech emotion recognition \cite{eyben2016geneva}. In order to obtain baseline results, we also used Logistic Regression and Random Forest. These are very powerful and widely used benchmark classifiers. The baseline results are presented in Figure \ref{Bls}. It can be noted from the Figure \ref{Bls} that SVM outperformed others. Therefore, we only mention the results using SVM beyond this experiment. 

\begin{figure*}[!ht]%
\centering
\begin{subfigure}{0.48\linewidth}
\includegraphics[trim=1.9cm 0cm 5cm 0.8cm,clip=true,width=\linewidth]{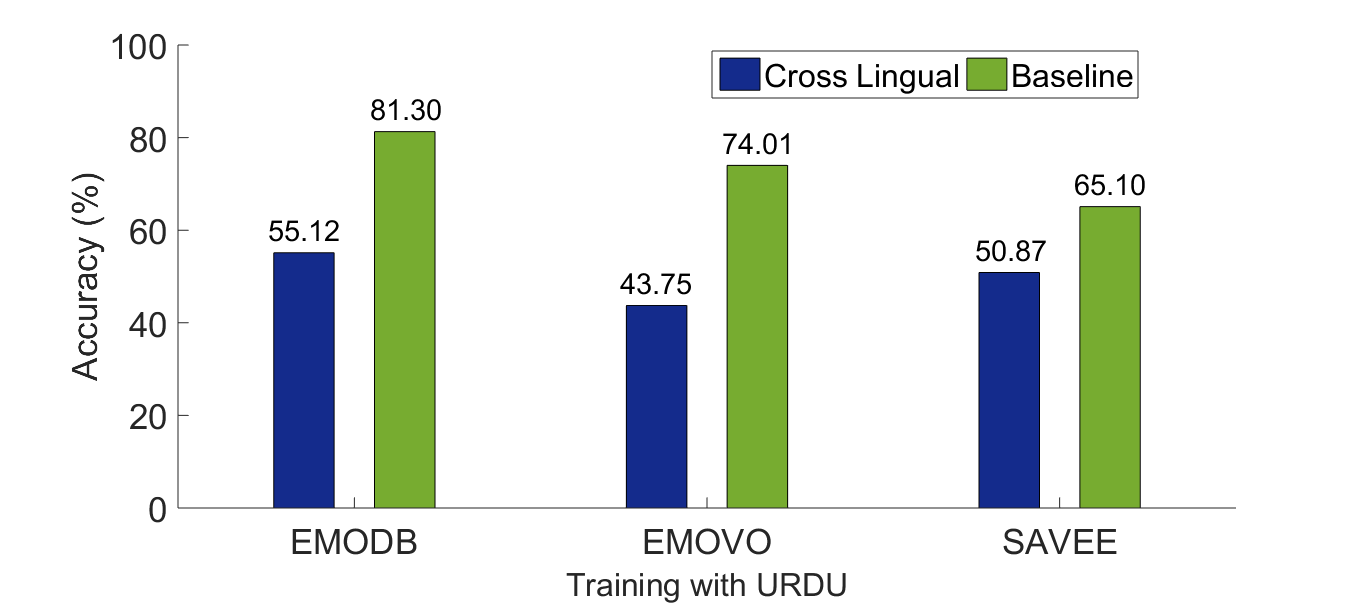}%
\captionsetup{justification=centering}
\caption{}%
\label{Urdu}%
\end{subfigure}
\begin{subfigure}{0.48\linewidth}
\includegraphics[trim=1.9cm 0cm 5cm 0.8cm,clip=true,width=\linewidth]{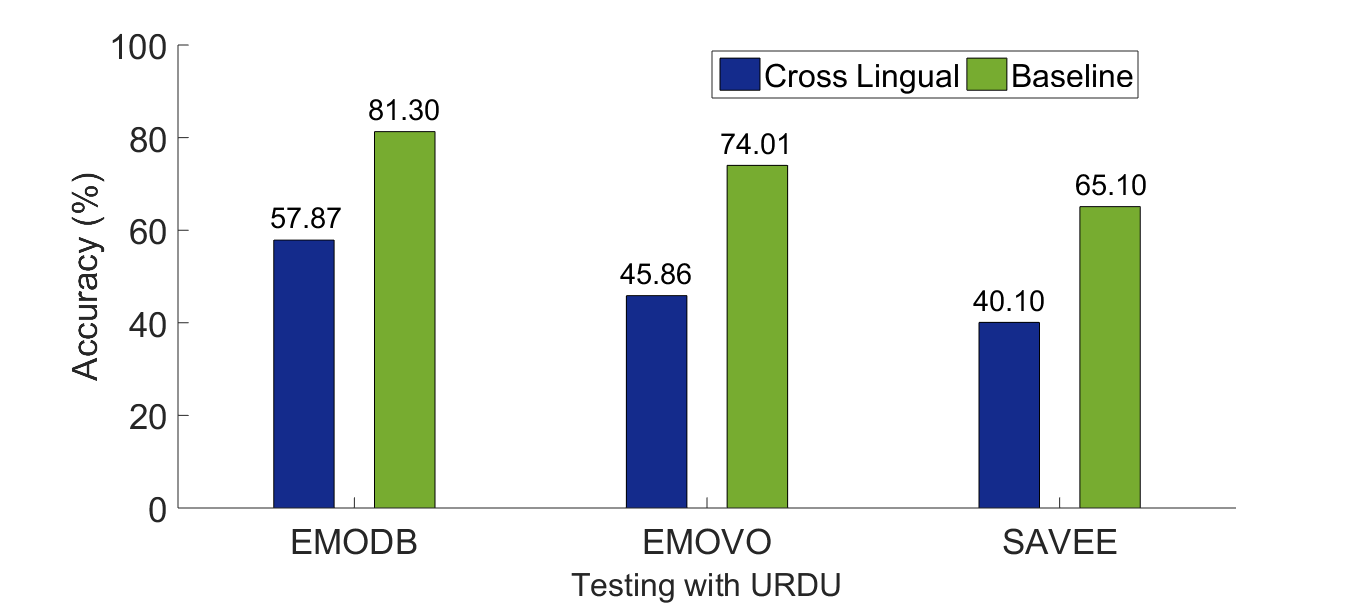}%
\captionsetup{justification=centering}
\caption{} %
\label{Western}%
\end{subfigure}%
\caption{Results for language training test using URDU data as training in (\ref{Urdu}) and (\ref{Western}) shows the training with Western languages.}
\label{Language}
\end{figure*}
\begin{figure*}[!t]%
\centering
\begin{subfigure}{0.48\linewidth}
\includegraphics[trim=1.9cm 0cm 4.2cm 0.8cm,clip=true,width=\linewidth]{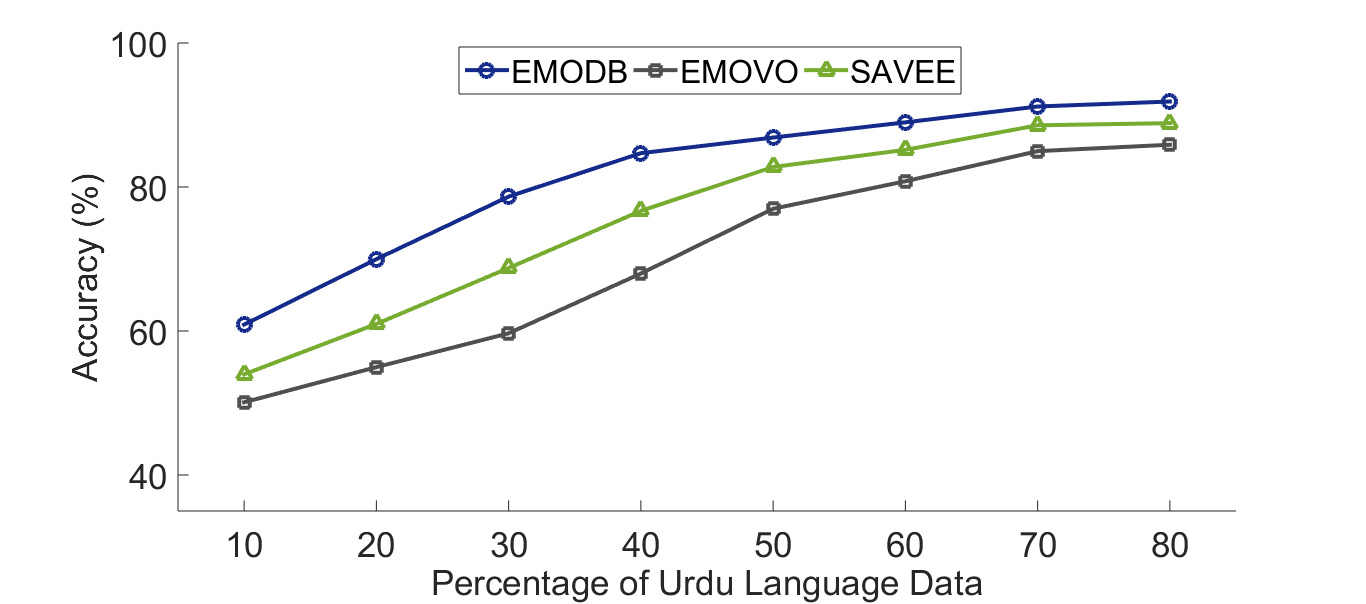}%
\captionsetup{justification=centering}
\caption{}%
\label{Ur}%
\end{subfigure}\hfill%
\begin{subfigure}{0.48\linewidth}
\includegraphics[trim=1.9cm 0cm 4.2cm 0.8cm,clip=true,width=\linewidth]{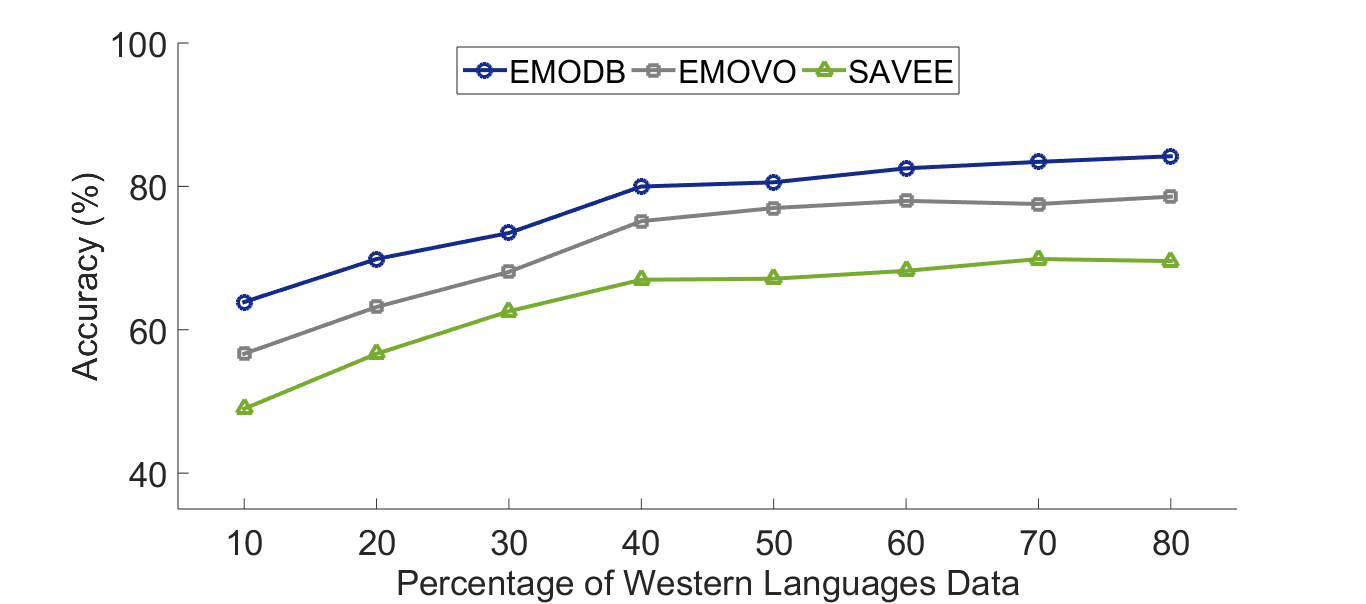}%
\captionsetup{justification=centering}
\caption{} %
\label{Wes}%
\end{subfigure}%
\caption{Impact of using a percentage of test date with training data. (\ref{Ur}) is for training with Western languages plus parentage of Urdu data and its inverse is shown in \ref{Wes}.}
\label{Percentage}
\end{figure*}

\subsection{Language Training Test}
In this experiment, we trained the model using data from one of the Western languages (English, Italian, and German) and testing is performed using Urdu data. Similarly, the inverse results are obtained by training with Urdu language data and testing the model using Western languages independently. Figure \ref{Language} highlights the trend of cross-lingual speech emotion recognition, which clearly shows that the accuracy for both of these scenarios drops significantly as compared to baseline results. This is mostly due to different studio conditions, labeling schemes, age groups, instruments used, and languages.  


\subsection{Percentage of Test Data}

In this experiment, we used percentage (10-80\%) of test data with training data. The model is trained using both Urdu and Western languages independently and tested by varying the percentage of testing language in training data. The results are presented in Figure \ref{Percentage}. It can be noted from the obtained results that the emotion classification accuracy across different languages is increased by incorporating the percentage of test language data. This means that the fraction of target data with training can help to achieve improved results even better than baseline accuracy. 

\subsection{Multi-language Training}

\begin{figure}[!ht]
\centering
\centerline{\includegraphics[trim=1cm 0cm 2.5cm 0.8cm,clip=true,width=.5\textwidth]{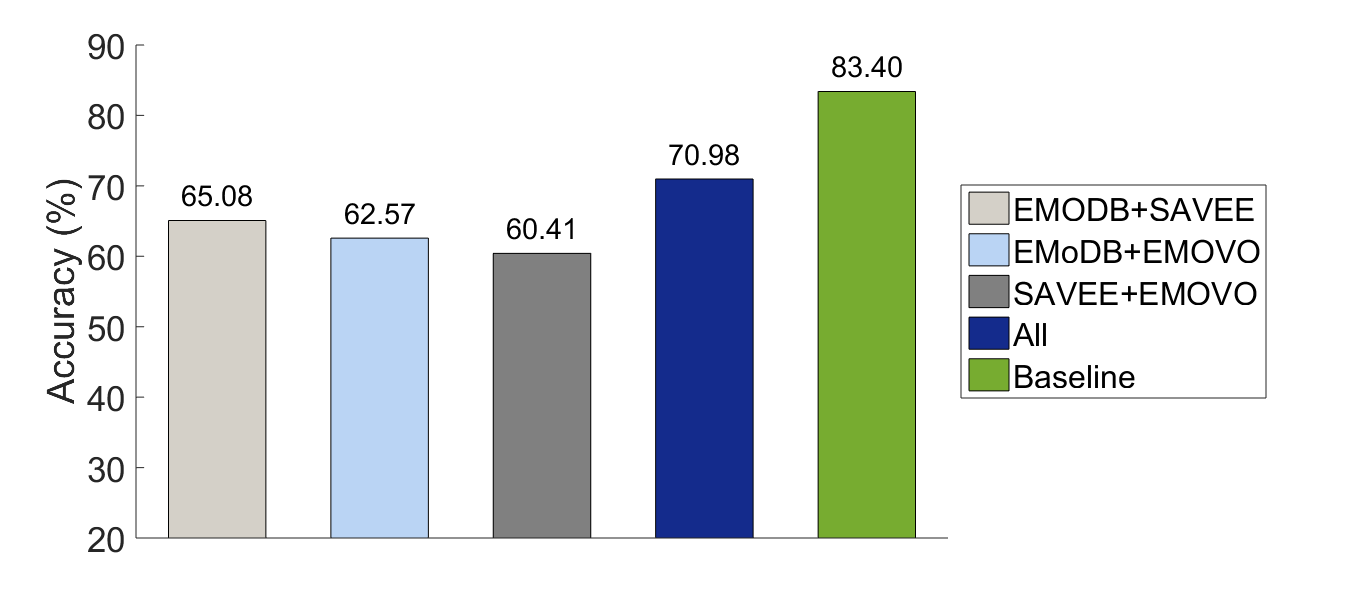}}
\caption{Results using different schemes for training the model against URDU data}
\label{Multi}
\end{figure}

In this experiment, we trained the model by fusing the data of Western languages to observe the improvement in the model's performance against Urdu language. We used different combinations of datasets and tested this setting for Urdu language. Figure \ref{Multi} shows the impact of multi-language training and compares the performance with baseline accuracy. It can be noted that the performance of emotion detection across languages can be increased using data from multiple languages. This is an important finding for real-time applications of cross-corpus emotion recognition.

\section{Discussion}
\label{dis}
We have presented different pilot experiments in an important field: cross-lingual emotion recognition for Urdu and Western languages. Significant improvement is achieved in emotion classification accuracy in few scenarios, which indicates that cross-lingual emotion recognition is possible even for Urdu language data collected from the spontaneous discussions of TV talk shows. However, it needs more efforts using more powerful cutting-edge machine learning or deep learning models to address this problem for minority languages like Urdu. We pointed out the following insights from this study. 

Figure \ref{Percentage} highlights an important practical aspect of our study that including a fraction of data from testing language into training data can help to achieve performance better than baseline results. Based on our experiments, augmenting above 30\% testing data in training can achieve accuracy even better than the baseline results. This is not only proved to be valid for Urdu language but also help for emotion recognition of Western languages while training the model with Urdu language data. These findings are very useful in real-time applications where target data is not available or very limited. 

From the experiments, multi-language training of model seems another practical example for dealing with minority languages like Urdu. This means that models' training with a wide range of languages can help to achieve accuracy comparable to the baseline. The performance of SVM (see Figure 4) on Urdu using three Western is a prime example of this scenario. When the model is trained using different combinations of two Western languages, the classification accuracy is much better than model's training with single Western language. Encouragingly, the accuracy is further improved by training SVM using data from three Western languages. However, the accuracy while training the model with three languages is lower than the baseline but not much poor as obtained when training is performed using a single Western language. 

The accuracy of SVM in the language training test in Figure 2 is much poor than the baseline results. The drop in performance against Urdu is different when the model is trained using different languages. Training on German data against Urdu gives better results as compared to other languages. This means that speech features in these two datasets have many similarities for emotion detection. The different studio conditions, language and age differences, and type of emotional corpus are the main reasons for the drop in accuracy. For instance, URDU data includes audio recordings from talk shows that contain real emotions whereas other three datasets comprise utterances in acted emotions recorded in studio conditions by professional actors. In addition, Urdu language data also have background noise in utterances. For practical applications, the issue of accuracy drop in language training experiment can be addressed significantly by previous two findings, i.e., either by training the model with multiple languages or by augmenting a portion of testing language data with training. 

\section{Conclusion}
\label{con}

This article has shown the fallbacks of current systems for speech emotion recognition using data of Urdu and Western languages. For this, we formulated the first spontaneous emotional URDU dataset and empirically evaluated cross-lingual emotion recognition for Urdu language using three Western languages. We have used support vector machine (SVM) for emotion recognition in intra- and inter-corpus scenarios and learned that the accuracy depends on conditions adopted for database formulation. When data of multiple languages are used for training, results for emotion detection is increased even for URDU dataset, which is highly dissimilar from other databases. Also, accuracy boosted when a small fraction of testing data is included in the training of the model with single corpus. These findings would be very helpful for designing a robust emotion recognition systems even for the languages having limited or no dataset. In our future work, we aim to further explore this problem by exploiting deep learning models for cross-lingual feature extraction and normalization techniques.

\section*{Acknowledgment}
We would like to thank Farwa Anees, Muhammad Usman, Muhammad Atif, and Farid Ullah Khan for assisting us in preparation of URDU dataset.

%



\end{document}